\documentclass[conference,a4paper]{IEEEtran}
\usepackage[noadjust]{cite}

\usepackage{times}
\usepackage{epsfig}
\usepackage{graphicx}
\usepackage[cmex10]{amsmath}
\usepackage{amssymb}
\usepackage{algorithmic}
\usepackage{array}
\usepackage{ragged2e}
\usepackage{mdwmath}
\usepackage{mdwtab}
\ifCLASSINFOpdf
\else
\fi
\begin{document}
%
\title{Stop-band Energy Constraint for Orthogonal Tunable Wavelet Units in Convolutional Neural Networks for Computer Vision problems}

\author{\IEEEauthorblockN{An D. Le\IEEEauthorrefmark{1}, Hung Nguyen\IEEEauthorrefmark{1}, Sungbal Seo\IEEEauthorrefmark{2}, You-Suk Bae\IEEEauthorrefmark{2} and Truong Q. Nguyen\IEEEauthorrefmark{1}}

\IEEEauthorblockA{\IEEEauthorrefmark{1}Jacobs School of Engineering, University of California San Diego, La Jolla, CA 92093, USA \\
\{d0le,hun004,tqn001\}@ucsd.edu\\
\IEEEauthorrefmark{2}Department of Computer Engineering, Tech University of Korea, Siheung 15073, South Korea\\
\{sungbal,ysbae\}@tukorea.ac.kr}}


\maketitle

\begin{abstract}
This work introduces a stop-band energy constraint for filters in orthogonal tunable wavelet units with a lattice structure, aimed at improving image classification and anomaly detection in CNNs, especially on texture-rich datasets. Integrated into ResNet-18, the method enhances convolution, pooling, and downsampling operations, yielding accuracy gains of 2.48\% on CIFAR-10 and 13.56\% on the Describable Textures dataset. Similar improvements are observed in ResNet-34. On the MVTec hazelnut anomaly detection task, the proposed method achieves competitive results in both segmentation and detection, outperforming existing approaches.
\end{abstract}


%
\IEEEpeerreviewmaketitle

\section{Introduction}
\label{sec:intro}
\begin{figure}[b!]
\begin{center}
\includegraphics[width=0.95\linewidth]{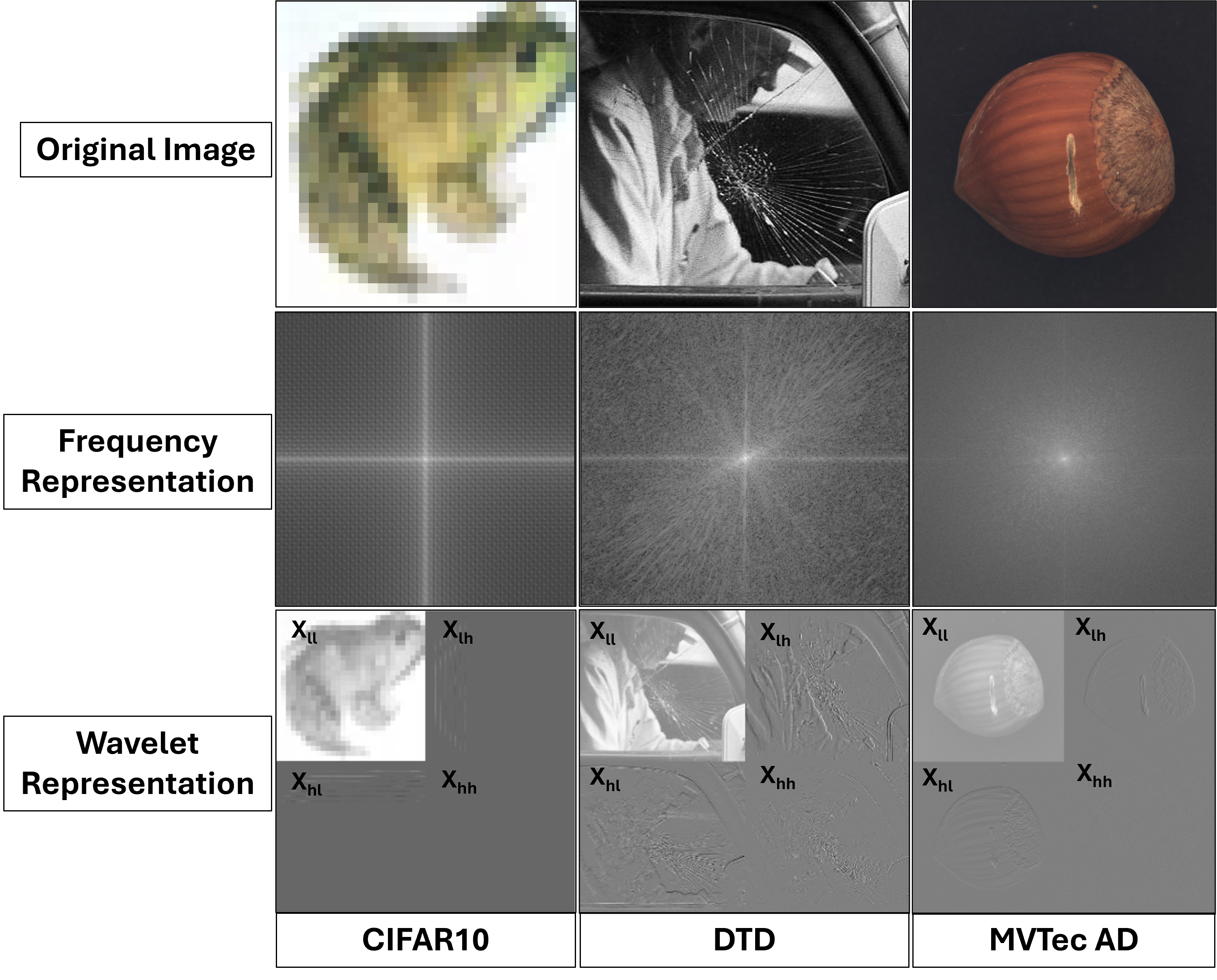}
\end{center}
\caption{From left to right, wavelet (Haar) and frequency representations of the samples from CIFAR10 (first column), DTD (second column), and MVTecAD (third column). The original images (top row) are shown with its frequency representation (middle row) and wavelet representation (bottom row). $\textbf{X}_{ll}$, $\textbf{X}_{lh}$, $\textbf{X}_{hl}$, and $\textbf{X}_{hh}$ show the coarse approximation and details wavelet representations.} 
\label{fig:Frequency_Wavelet_Analysis}
\end{figure}
Convolutional Neural Networks (CNNs) are widely used in various computer vision applications, including medical imaging \cite{wavelet_rp, uwu_ILM-ERM}, smart agriculture \cite{AlertTrap_journal}, and air-quality monitoring \cite{sheaf_theory, PM2_5_CDL}. Max pooling, a key component in CNN architectures such as ResNets \cite{Resnet}, emphasizes dominant features but discards fine details, leading to aliasing artifacts \cite{Alias_CNN}. While frequency-based methods \cite{Spectral_Pooling, diffstride} focus on low-frequency components, wavelet-based models like WaveCNet \cite{WaveCNET} predominantly use low-pass filters. However, models such as Wavelet-Attention CNNs \cite{Wavelet-AttentionCNN} incorporates both coarse and fine-grained details, which is crucial for high-resolution image processing.

As illustrated in Fig. \ref{fig:Frequency_Wavelet_Analysis}, the CIFAR-10 dataset \cite{CIFAR10} primarily contains low-frequency information, whereas MVTecAD \cite{MVTecAD_1,MVTecAD_2} and DTD \cite{DTD} distribute features across both low- and high-frequency regions. In the "cracked" DTD sample shown in the third column of Fig. \ref{fig:Frequency_Wavelet_Analysis}, the low-pass component $\textbf{X}_{ll}$ retains minimal texture information, while the high-pass components $\textbf{X}_{hl}$, $\textbf{X}_{lh}$, and $\textbf{X}_{hh}$ capture the distinctive characteristics. This highlights the necessity of preserving both high-pass and low-pass information in CNNs.

Previous studies \cite{rp_clinical, ilm-erm_clinical} explored methods to retain full image information by leveraging wavelet decomposition and perfect reconstruction, demonstrating performance gains in clinical settings. The introduction of tunable wavelet filters in \cite{OrthLatt_UwU, bioruwu} further improved CNN processing, particularly for images with significant high-frequency content. However, the flexibility of these filters resulted in an unclear distinction between high-pass and low-pass behaviors.

To address this, we propose a Stopband Energy Constraint (SBE) to enforce distinct low-pass and high-pass characteristics in tunable wavelet filters. The benefits of the proposed SBE are task-dependent and particularly pronounced in fine-detail preservation scenarios. The proposed loss function is applied to the Orthogonal Lattice Structure Universal Wavelet Unit (OrthLatt-UwU) \cite{OrthLatt_UwU} and evaluated using ResNet architectures on CIFAR-10 and DTD, where it demonstrates clear advantages. Additionally, models trained on DTD serve as feature extractors in the CFLOW-AD \cite{CFLOW_Gudovskiy_2022_WACV} anomaly detection pipeline, tested on the hazelnut category in the MVTecAD dataset \cite{MVTecAD_1,MVTecAD_2}. In summary:

\begin{itemize}
    \item We introduce a Stopband Energy Constraint to enforce clear low-pass and high-pass characteristics in wavelet-tunable filters.
    \item We apply the proposed method to train OrthLatt-UwU within ResNet architectures on CIFAR-10 and DTD, achieving improved classification performance.
    \item The proposed units are integrated into the CFLOW-AD anomaly detection model and evaluated on MVTecAD.
\end{itemize}

\section{Related Works}
\begin{figure}[b]
\begin{center}
\includegraphics[width=\linewidth]{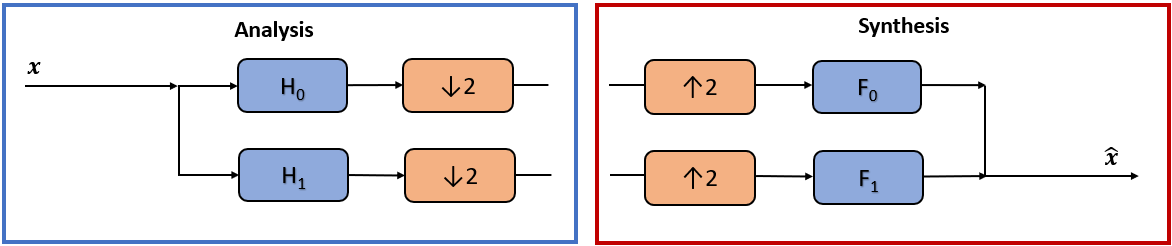}
\end{center}
\caption{Two-channel filter bank architecture.}
\label{fig:filterbank}
\end{figure}
\label{sec:relatedwork}
Max pooling reduces feature map resolution by selecting maximum values, preserving dominant features \cite{Feature-Pooling_in_Visual-Recognition, MaxPool}, but can introduce aliasing and structural degradation \cite{Alias_CNN, WaveCNET}. Alternatives like spectral pooling \cite{Spectral_Pooling} retain low-frequency content but rely on fixed hyperparameters, while DiffStride \cite{diffstride} enables differentiable stride learning yet still loses high-frequency detail. Wavelet-based methods \cite{Mallat, Wavelets_and_filter_banks} offer multiscale analysis, though most focus only on approximation components \cite{WaveCNET}. Recent works about trainable filters \cite{AnLe, OrthLatt_UwU, bioruwu} have introduced learning-based wavelet pooling, but lacked clear frequency separation. To address this, we introduce a stop-band energy constraint that enforces distinct low-pass and high-pass behavior, improving detail preservation for texture-rich images.
\begin{figure}[!t]
\begin{center}
\includegraphics[width=\linewidth]{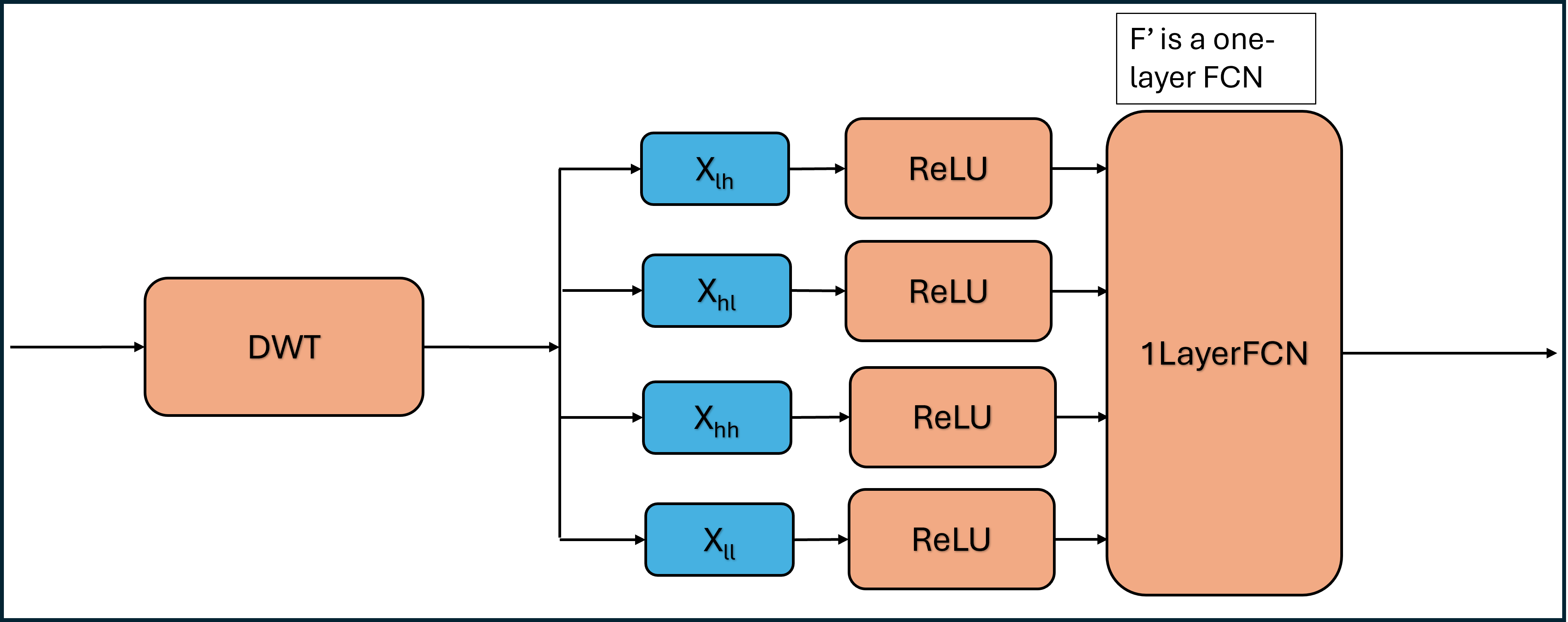}
\end{center}
\caption{\justifying{Diagram of low-pass and high-pass component implementation. The signal goes from left to right. The results from DWT go to ReLU functions to become the inputs of a one-layer FCN. Because an FCN can take inputs of arbitrary sizes, the one-layer FCN can read the decomposed components and finetune the trainable coefficients to optimally combine the decomposed components. The fine-tuned one-layer FCN combines the inputs to find the optimal feature map.}}
\label{fig:OneLayerFCN}
\end{figure}

\begin{figure}[!b]
\begin{center}
\includegraphics[width=\linewidth]{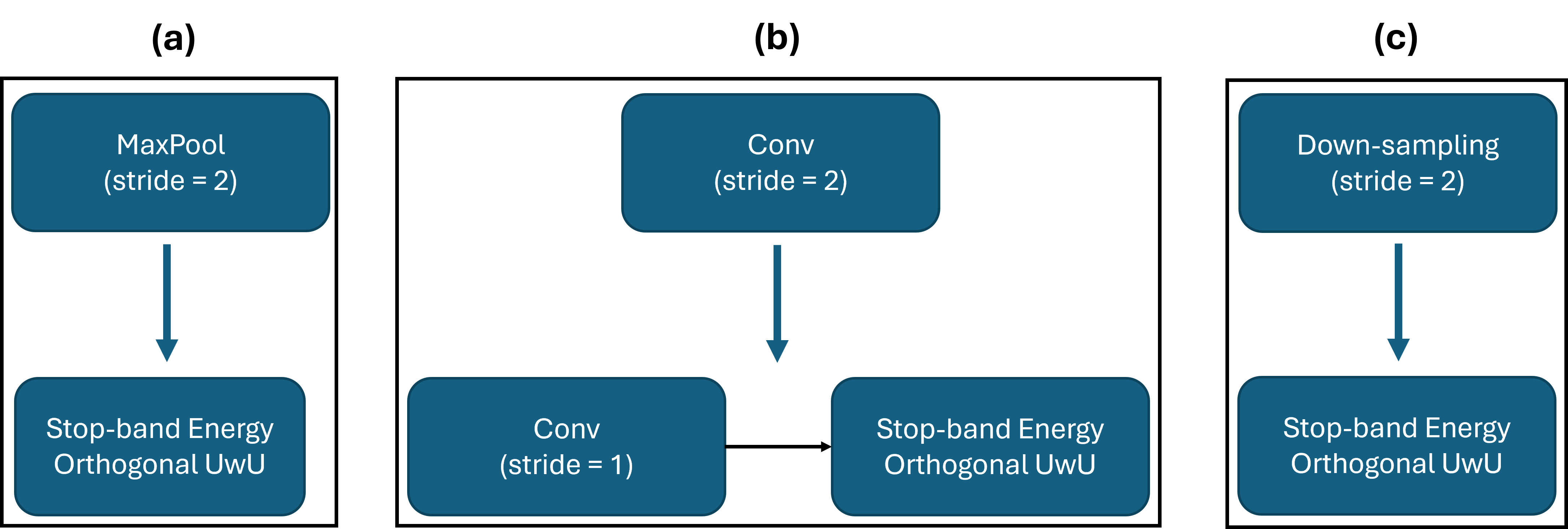}
\end{center}
\caption{Implementation of the proposed unit in CNN architecture, replacing max-pool (a), stride-convolution (b), and downsampling (c) functions.}
\label{fig:UwU2}
\end{figure}
\section{Proposed Method}
\label{sec:method}
\subsection{Stop-band Energy constraint for Tunable Orthogonal Wavelet Filters}
\begin{figure*}[!t]
\begin{center}
\includegraphics[width=0.9\linewidth]{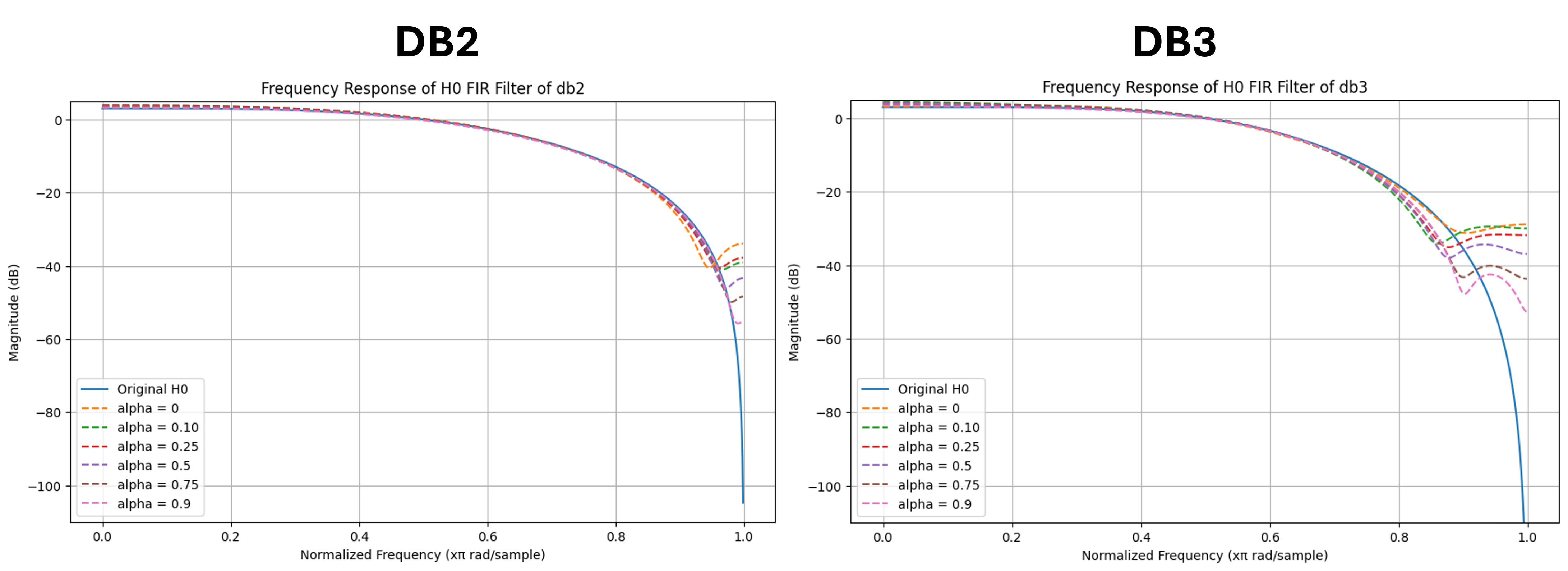}
\end{center}
\caption{Frequency Response Analysis of tuned DB2 (left) and DB3 (right) with the applied stop-band energy constraint. The solid blue line denotes the original H0 filter, while the dashed lines represent tuned filters with varying $\alpha$ values from 0 to 0.9.} 
\label{fig:Frequency_Response_Analysis}
\end{figure*}
The stop-band Energy constraint is to enforce low-pass and high-pass behaviors in $\textbf{h}_0$ and $\textbf{h}_1$, respectively, in the filter banks of tunable wavelet units to improve feature extraction. The filter bank structure is shown in Fig. \ref{fig:filterbank}. In Fig. \ref{fig:filterbank}, the analysis, shown in the blue rectangle box, and synthesis, shown in the red rectangle box, parts of the filter bank have the function of decomposing and reconstructing signals, respectively. $H_{0}$ and $H_{1}$ are, correspondingly, low-pass and high-pass filters for the analysis part of the filter bank; whereas $F_{0}$ and $F_{1}$ are, respectively, low-pass and high-pass filters for the synthesis part of the filter bank. With $N$ taps, $H_{0}$ and $H_{1}$ have $\textbf{h}_{0}=[h_0(0), h_0(1),..., h_0(N-1)]$ and $\textbf{h}_{1}=[h_1(0), h_1(1),..., h_1(N-1)]$ as their coefficients, respectively. For $H_0$, the stop-band energy $E_{sb}$ is defined as follows:
\begin{equation}
\label{eq:StopbandEnergy_H0}
E_{sb} = \frac{1}{2\pi}\int_{w_s}^{\pi} |H_0(e^{jw})|^2 \,dw,
\end{equation}
in which $|H_0(e^{jw})|^2$ is the energy of $H_0$ at $w$ and $w_s$ is the stop-band frequency. For $H_1$, the same formula is used with the integral part from $0$ to $w_s$. In this work, $w_s$ is set to be $0.6\pi$ and $0.4\pi$ for $H_0$ and $H_1$, respectively. Nevertheless, as the stop-band energy constraint is applied on orthogonal filter bank in this work, only the stopband of $H_0$ is needed to due to the relationship between $H_0$ and $H_1$. For the implementation of numerical integration, the range from \( 0 \) to \( \pi \) is divided into \( K \) smaller intervals to approximate the integral by summing contributions over these sub-intervals. Hence, we have the numerical implementation $E_{sbn}$ of $E_{sb}$ as follows:
\begin{equation}
\label{eq:NormalizedStopbandEnergy_Implementation}
E_{sbn} = \frac{1}{2K}\sum_{i=w_{si}}^{K}|H_0(e^{jw_{i}})|^2,
\end{equation}
where $w_{si}$ is the approximated frequency sub-interval for the stop-band frequency $w_s$ defined as $w_{si} = round(K*\frac{w_s}{\pi})$ and $w_i$ is a frequency sub-interval defined as $w_i = i*\frac{\pi}{K}$ for $i = 0,1,..K$. In this study, $K$ is set to be $500$, and for higher-precision, a higher value can be used. Finally, we normalize the approximated stop-band energy with the total energy of $H_0$. Therefore, we have the normalized stop-band energy loss function $L_{SBE}$ defined as follows:
\begin{equation}
\label{eq:StopbandEnergy_H0_loss}
L_{SBE} = \frac{E_{sbn}}{E_{total}}=\frac{1}{2K}\frac{\sum_{i=w_{si}}^{K}|H_0(e^{jw_{i}})|^2}{\sum_{k=0}^{N-1} {h_0[k]}^2}.
\end{equation}

From (\ref{eq:StopbandEnergy_H0_loss}), stop-band energy constraint is implemented to train the filter bank analysis. The effect of the constraint on the coefficients can be controlled by multiplying $L_{SBE}$ with a factor $\alpha$. In our image classification study using Cross-Entropy $L_{CE}$. A higher $\alpha$ strengthens the stop-band constraint, while having $\alpha = 0$ means no stop-band energy constraint being applied. The total loss function $L_{total}$ is expressed as follows:
\begin{equation}
\label{eq:Totalloss}
L_{total} = (1-\alpha)L_{CE} + \alpha L_{SBE}.
\end{equation}

\subsection{2D Implementation}
With \( L_{SBE} \), the filter coefficients \( \textbf{h}_0 \) and \( \textbf{h}_1 \) are regularized during training. From these coefficients, the high-pass and low-pass filter matrices \( \textbf{H} \) and \( \textbf{L} \) are derived to compute the approximation component \( X_{ll} \) and the detail components \( X_{lh} \), \( X_{hl} \), and \( X_{hh} \). The matrix \( \textbf{L} \) is computed as follows:
\begin{equation}
\label{eq:L}
 \textbf{L} = \textbf{D}\widehat{\textbf{H}},
\end{equation}
where \( \textbf{D} \) represents the downsampling matrix, and \( \widehat{\textbf{H}} \) is a Toeplitz matrix constructed from the filter coefficients of \( \textbf{H}_{0}(z) \). The matrix \( \textbf{H} \) follows a similar structure to \( \textbf{L} \) but uses the filter coefficients of \( \textbf{H}_{0}(z^{-1}) \). With \( \textbf{H} \) and \( \textbf{L} \), the components \( X_{ll} \), \( X_{lh} \), \( X_{hl} \), and \( X_{hh} \) are computed as follows:
\begin{equation}
\label{eq:decomposition}
\begin{aligned}
 \textbf{X}_{ll} &=  \textbf{L}\textbf{X}\textbf{L}^T, &
 \textbf{X}_{lh} &=  \textbf{H}\textbf{X}\textbf{L}^T,\\
 \textbf{X}_{hl} &=  \textbf{L}\textbf{X}\textbf{H}^T, &
 \textbf{X}_{hh} &=  \textbf{H}\textbf{X}\textbf{H}^T,
\end{aligned}
\end{equation}

\subsection{Implementation in CNN architectures}
The proposed units were integrated into ResNet family architectures. For downsampling and pooling layers, the UwU is followed by a one-layer fully connected network (FCN), as illustrated in Fig. \ref{fig:OneLayerFCN}. Additionally, the stride-2 convolution is replaced with a non-stride convolution block, followed by the proposed UwU, as shown in Fig. \ref{fig:UwU2}.

\section{Experiments and Results}
\label{sec:exandresults}
\begin{table}[b!]
\caption{\justifying{Accuracy of SBE-OrthLatt-UwU in ResNet18 with different number of taps and initialization-wavelet types trained on DTD and CIFAR10.\\}}
\centering
\scalebox{0.8}{
\begin{tabular}{|c|c|c|c|c|c|c|}
\hline
\multicolumn{1}{|c|}{} & \multicolumn{3}{c|}{\textbf{CIFAR10}} & \multicolumn{3}{c|}{\textbf{DTD}} \\ \hline
\multicolumn{1}{|c|}{ResNet18} & \multicolumn{3}{c|}{92.44$\%$} & \multicolumn{3}{c|}{33.99$\%$} \\ \hline
\multicolumn{1}{|c|}{Wavelet} & \multicolumn{1}{c|}{$\alpha:0.25$} & \multicolumn{1}{c|}{$\alpha:0.5$}& \multicolumn{1}{c|}{$\alpha:0.75$}& \multicolumn{1}{c|}{$\alpha:0.25$}& \multicolumn{1}{c|}{$\alpha:0.5$}& \multicolumn{1}{c|}{$\alpha:0.75$}\\ \hline
\multicolumn{1}{|c|}{Haar} & \multicolumn{1}{c|}{$94.54$} & \multicolumn{1}{c|}{$94.83$}& \multicolumn{1}{c|}{$94.11$}& \multicolumn{1}{c|}{$41.86$}& \multicolumn{1}{c|}{$45.69$}& \multicolumn{1}{c|}{$47.55$}\\ \hline
\multicolumn{1}{|c|}{DB2} & \multicolumn{1}{c|}{$94.92$} & \multicolumn{1}{c|}{$94.54$}& \multicolumn{1}{c|}{$94.07$}& \multicolumn{1}{c|}{$42.98$}& \multicolumn{1}{c|}{$43.94$}& \multicolumn{1}{c|}{$46.65$}\\ \hline
\multicolumn{1}{|c|}{DB3} & \multicolumn{1}{c|}{$94.60$} & \multicolumn{1}{c|}{$94.46$}& \multicolumn{1}{c|}{$94.25$}& \multicolumn{1}{c|}{$38.40$}& \multicolumn{1}{c|}{$44.10$}& \multicolumn{1}{c|}{$46.28$}\\ \hline
\multicolumn{1}{|c|}{DB4} & \multicolumn{1}{c|}{$93.85$} & \multicolumn{1}{c|}{$94.48$}& \multicolumn{1}{c|}{$94.50$}& \multicolumn{1}{c|}{$39.04$}& \multicolumn{1}{c|}{$39.10$}& \multicolumn{1}{c|}{$46.54$}\\ \hline
\end{tabular}}
\label{table:Cifar10_DTD}
\end{table}
This section applies the proposed stop-band energy constraint to OrthLatt-UwUs. First, its impact on frequency response is analyzed. Then, it is implemented in ResNet18 (tested on CIFAR10 and DTD with Haar, db2, db3, and db4) and ResNet34 (with Haar and db2). Finally, its effectiveness is evaluated in the CFLOW-AD pipeline for anomaly detection on the hazelnut class in the MVTecAD dataset.
\begin{table}[b!]
\caption{\justifying{Accuracy of SBE-OrthLatt-UwU in ResNet18 with other wavelet based approaches on CIFAR10.\\}}
\centering
\scalebox{0.68}{
\begin{tabular}{|c|c|c|c|c|}
\hline
\multicolumn{1}{|c|}{\textbf{Wavelet}} & \multicolumn{4}{c|}{\textbf{Accuracy($\%$)}} \\ \hline
\multicolumn{1}{|c|}{None (Baseline)} & \multicolumn{4}{c|}{92.44} \\ \hline
\multicolumn{1}{|c|}{} & \multicolumn{1}{c|}{OrthLatt UwU} & \multicolumn{1}{c|}{Ours} & \multicolumn{1}{|c|}{PR-Relax UwU} & \multicolumn{1}{c|}{WaveCNet} \\ \hline
\multicolumn{1}{|c|}{2-tap Haar} & \multicolumn{1}{c|}{94.97} & \multicolumn{1}{c|}{94.83 ($\alpha=0.5$)} & \multicolumn{1}{c|}{94.97} & \multicolumn{1}{c|}{94.76} \\ \hline
\multicolumn{1}{|c|}{4-tap DB2} & \multicolumn{1}{c|}{95.05} & \multicolumn{1}{c|}{94.92 ($\alpha=0.25$)} & \multicolumn{1}{c|}{94.66} &  \multicolumn{1}{c|}{94.93} \\ \hline
\multicolumn{1}{|c|}{6-tap DB3} & \multicolumn{1}{c|}{95.03} & \multicolumn{1}{c|}{94.60 ($\alpha=0.25$)} & \multicolumn{1}{c|}{93.37} & \multicolumn{1}{c|}{94.56} \\ \hline
\multicolumn{1}{|c|}{8-tap DB4} & \multicolumn{1}{c|}{95.11} & \multicolumn{1}{c|}{94.50 ($\alpha=0.75$)} & \multicolumn{1}{c|}{94.21} & \multicolumn{1}{c|}{93.81} \\ \hline
\end{tabular}}
\label{table:Cifar10_Wavelet}
\end{table}
\subsection{Stop-band Energy Constraint Effect: Frequency Response Analysis}
This section investigates the impact of the stop-band energy constraint on filter behavior. In this experiment, the constraint was applied to OrthLatt-UwU, replacing the first pooling unit in a ResNet model trained on CIFAR-10. The effect on frequency selectivity was evaluated by varying \(\alpha\) values (\(0, 0.1, 0.25, 0.5, 0.75, 0.9\)), where \(\alpha = 0\) indicates no constraint. The \( h_0 \) coefficients of db2 and db3 wavelets were used for initialization. As shown in Fig. \ref{fig:Frequency_Response_Analysis}, higher \(\alpha\) values (e.g., 0.75, 0.9) result in stronger stopband attenuation, enhancing frequency selectivity, while lower \(\alpha\) values (e.g., 0, 0.10) exhibit more distorted low-pass behavior. This confirms that increasing \(\alpha\) enforces low-pass characteristics and improves filter performance by preserving both low-pass and high-pass features.
\subsection{Image Classification: CIFAR10 and DTD}
This section evaluates the performance of the stop-band energy (SBE) constraint on OrthLatt-UwU units, referred to as SBE-OrthLatt-UwU, implemented in ResNet architectures on CIFAR-10 and DTD. CIFAR-10 \cite{CIFAR10} consists of 60,000 low-resolution (32×32) color images across 10 classes, with 50,000 for training and 10,000 for testing. In contrast, DTD \cite{DTD} (Describable Textures Dataset) contains 5,640 high-resolution images across 47 categories, emphasizing rich textural details and high-frequency information. The proposed constraint was primarily tested on ResNet18 and compared against WaveCNet \cite{WaveCNET}, PR-Relax UwU \cite{AnLe}, and OrthLatt-UwU \cite{OrthLatt_UwU}. The best-performing SBE-OrthLatt-UwU model from each experiment was further compared with reported results from other methods. Additionally, experiments were extended to ResNet34 to assess whether the observed improvements in ResNet18 remained consistent in deeper models.
\begin{table}[t!]
\caption{\justifying{Accuracy of SBE-OrthLatt-UwU in ResNet18 with other wavelet based approaches on DTD.\\}}
\centering
\scalebox{0.78}{
\begin{tabular}{|c|c|c|c|c|}
\hline
\multicolumn{1}{|c|}{\textbf{Wavelet}} & \multicolumn{4}{|c|}{\textbf{Accuracy($\%$)}} \\ \hline
\multicolumn{1}{|c|}{None (Baseline)} & \multicolumn{4}{|c|}{33.99} \\ \hline
\multicolumn{1}{|c|}{} & \multicolumn{1}{|c|}{OrthLatt UwU} & \multicolumn{1}{|c|}{Ours} & \multicolumn{1}{|c|}{PR-Relax UwU} &\multicolumn{1}{|c|}{WaveCNet} \\ \hline
\multicolumn{1}{|c|}{2-tap Haar} & \multicolumn{1}{|c|}{40.37} & \multicolumn{1}{|c|}{47.55} & \multicolumn{1}{|c|}{43.67} & \multicolumn{1}{|c|}{25.53} \\ \hline
\multicolumn{1}{|c|}{4-tap DB2} & \multicolumn{1}{|c|}{43.51} & \multicolumn{1}{|c|}{46.65} & \multicolumn{1}{|c|}{40.69} & \multicolumn{1}{|c|}{23.62} \\ \hline
\multicolumn{1}{|c|}{6-tap DB3} & \multicolumn{1}{|c|}{40.59} & \multicolumn{1}{|c|}{46.28} & \multicolumn{1}{|c|}{35.90} & \multicolumn{1}{|c|}{24.36} \\ \hline
\multicolumn{1}{|c|}{8-tap DB4} & \multicolumn{1}{|c|}{40.80} & \multicolumn{1}{|c|}{46.54} & \multicolumn{1}{|c|}{39.79} & \multicolumn{1}{|c|}{26.91} \\ \hline
\end{tabular}}
\label{table:DTD_Wavelet}
\end{table}
\subsubsection{ResNet18}
\begin{table}[t!]
\caption{\justifying{Accuracy of the best SBE-OrthLatt-UwU in ResNet18 model compared to other approaches with the same architecture on CIFAR10.\\}}
\centering
\scalebox{0.9}{
\begin{tabular}{cc}
\hline
\multicolumn{1}{|l|}{\textbf{Models}} & \multicolumn{1}{|c|}{\textbf{Accuracy($\%$)}} \\ \hline
\multicolumn{1}{|l|}{baseline-ResNet18} & \multicolumn{1}{|c|}{92.44} \\
\hline
\multicolumn{1}{|l|}{SpectralPool-ResNet18\cite{Spectral_Pooling}} & \multicolumn{1}{|c|}{92.50 (+0.06)} \\
\hline
\multicolumn{1}{|l|}{DiffStride-ResNet18\cite{diffstride}} & \multicolumn{1}{|c|}{92.90 (+0.46)} \\
\hline
\multicolumn{1}{|l|}{WA-CNN ResNet18\cite{Wavelet-AttentionCNN}} & \multicolumn{1}{|c|}{92.57 (+0.13)} \\
\hline
\multicolumn{1}{|l|}{WaveCNet ResNet18 DB2\cite{WaveCNET}} & \multicolumn{1}{|c|}{94.93 (+2.49)} \\
\hline
\multicolumn{1}{|l|}{PR-relax UwU (Haar) ResNet18\cite{AnLe}} & \multicolumn{1}{|c|}{94.97 (+2.52)} \\ \hline
\multicolumn{1}{|l|}{OrthLatt-UwU (DB4) ResNet18 \cite{OrthLatt_UwU}} & \multicolumn{1}{|c|}{95.11 (+2.67)} \\ \hline
\multicolumn{1}{|l|}{SBE-OrthLatt-UwU (DB2) ResNet18 (ours)} & \multicolumn{1}{|c|}{94.92 (+2.48)} \\ \hline
\end{tabular}}
\label{table:BestCIFAR10}
\end{table}
SBE-OrthLatt-UwU with \(\alpha\) values of 0.25, 0.5, and 0.75 was implemented in ResNet18 and tested on CIFAR10 and DTD, representing datasets with low-resolution, low-pass features and high-resolution, high-pass details, respectively. The model's coefficients were initialized with Haar, DB2, DB3, and DB4 wavelets for 2-tap, 4-tap, 6-tap, and 8-tap cases. Table \ref{table:Cifar10_DTD} shows that across all wavelet initializations and \(\alpha\) values, SBE-OrthLatt-UwU ResNet18 outperformed the baseline ResNet18. For DTD, higher \(\alpha\) values consistently led to better performance across all wavelet cases.

The best-performing SBE-OrthLatt-UwU ResNet18 models were compared with WaveCNet \cite{WaveCNET}, OrthLatt-UwU \cite{OrthLatt_UwU}, and PR-Relax UwU \cite{AnLe}. On CIFAR10, SBE-OrthLatt-UwU performs comparably to other wavelet-based methods (Table~\ref{table:Cifar10_Wavelet}) and outperforms spectrum-based approaches (Table~\ref{table:BestCIFAR10}). On DTD, it achieves the highest accuracy (Table~\ref{table:DTD_Wavelet}), especially with Haar and DB2 at \(\alpha = 0.75\). While its accuracy on CIFAR10 is slightly lower than OrthLatt-UwU, it clearly surpasses both the baseline and OrthLatt-UwU on fine-detail tasks. The average inference times are 0.021s (baseline), 0.0312s (WaveCNet), 0.0262s (OrthLatt-UwU), and 0.053s (SBE-OrthLatt-UwU), indicating a reasonable trade-off.

\subsubsection{Extended Study with ResNet34}
\begin{table}[t!]
\caption{\justifying{Accuracy of SBE-OrthLatt-UwU with \(\alpha\) values of 0.25 (left), 0.5 (middle), and 0.75 (right), along with OrthLatt-UwU and WaveCNet, evaluated on ResNet34 for DTD and CIFAR10 with different tap numbers and wavelets.\\}}
\centering
\scalebox{0.8}{
\begin{tabular}{cccccc}
\hline
\multicolumn{1}{|c|}{\textbf{Wavelet}} & \multicolumn{5}{|c|}{\textbf{DTD}} \\ \hline
\multicolumn{1}{|c|}{None (Baseline)} & \multicolumn{5}{|c|}{24.47$\%$} \\ \hline
\multicolumn{1}{|c|}{} & \multicolumn{1}{|c|}{OrthLatt-UwU} & \multicolumn{3}{|c|}{Stop-band Energy UwU (Ours)} & \multicolumn{1}{|c|}{WaveCNet} \\ \hline
\multicolumn{1}{|c|}{Haar} & \multicolumn{1}{|c|}{41.49$\%$} & \multicolumn{1}{|c|}{36.70$\%$} & \multicolumn{1}{|c|}{41.49$\%$}& \multicolumn{1}{|c|}{43.14$\%$}&\multicolumn{1}{|c|}{32.39$\%$} \\ \hline
\multicolumn{1}{|c|}{DB2} & \multicolumn{1}{|c|}{38.88$\%$} & \multicolumn{1}{|c|}{36.65$\%$}& \multicolumn{1}{|c|}{43.03$\%$}& \multicolumn{1}{|c|}{45.00$\%$} & \multicolumn{1}{|c|}{32.23$\%$} \\ \hline\hline
\multicolumn{1}{|c|}{\textbf{Wavelet}} & \multicolumn{5}{|c|}{\textbf{CIFAR10}}\\ \hline
\multicolumn{1}{|c|}{None (Baseline)} & \multicolumn{5}{|c|}{94.33$\%$}\\ \hline
\multicolumn{1}{|c|}{} & \multicolumn{1}{|c|}{OrthLatt-UwU} & \multicolumn{3}{|c|}{Stop-band Energy UwU (Ours)} & \multicolumn{1}{|c|}{WaveCNet} \\ \hline
\multicolumn{1}{|c|}{Haar} & \multicolumn{1}{|c|}{95.44$\%$} & \multicolumn{3}{|c|}{94.71$\%$ ($\alpha = 0.5$)} & \multicolumn{1}{|c|}{95.07$\%$}\\ \hline
\multicolumn{1}{|c|}{DB2} & \multicolumn{1}{|c|}{95.61$\%$} & \multicolumn{3}{|c|}{94.63$\%$ ($\alpha = 0.75$)} & \multicolumn{1}{|c|}{95.12$\%$}\\ \hline
\end{tabular}}
\label{table:DTD_CIFAR10_ResNet34}
\end{table}
In this section, SBE-OrthLatt-UwU with \(\alpha\) values of 0.25, 0.5, and 0.75, initialized with Haar and DB2 wavelets, was implemented in ResNet34, tested on CIFAR10 and DTD, and compared against baseline ResNet34, WaveCNet, and OrthLatt-UwU. As shown in Table \ref{table:DTD_CIFAR10_ResNet34}, while SBE-OrthLatt-UwU achieves comparable performance on CIFAR10, it consistently outperforms other methods across all wavelet types at \(\alpha = 0.75\) in DTD.

\subsection{Anomaly Detection: MVTecAD}
\begin{figure}[!t]
\begin{center}
\includegraphics[width=0.9\linewidth]{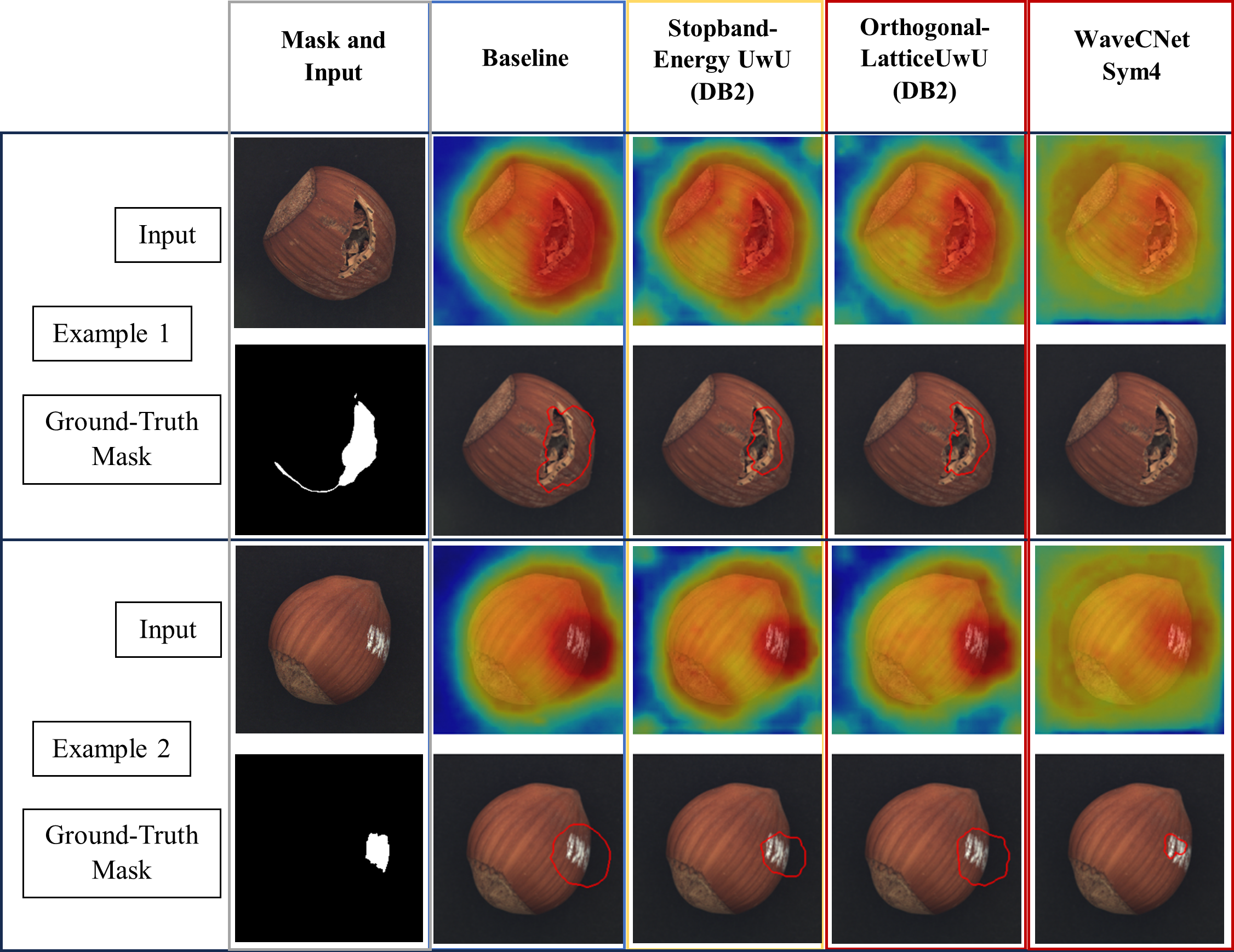}
\end{center}
\caption{\justifying{Anomaly detection on hazelnut objects from the MVTec AD dataset with two examples. From left to right: the first column presents the mask and input image, while the second to fifth columns display heatmaps and defect segmentation results from the baseline, Stopband-Energy UwU DB2, Orthogonal-Lattice UwU DB2, and WaveCNet Sym4, respectively.}} 
\label{fig:MVTecAD_hazelnut}
\end{figure}

In this experiment, SBE-OrthLatt-UwU ResNet18 (DB2, \(\alpha = 0.75\)) trained on DTD was used as the encoder in the CFLOW-AD pipeline \cite{CFLOW_Gudovskiy_2022_WACV} for anomaly detection on MVTecAD hazelnut images \cite{MVTecAD_1,MVTecAD_2}. As shown in Table~\ref{table:MVTecAD_hazelnut}, it achieved the best detection AUROC (\(94.46\%\)) and second-best segmentation AUROC (\(96.99\%\)), compared to the baseline, OrthLatt-UwU (\(97.15\%\) segmentation), and WaveCNet. Fig.~\ref{fig:MVTecAD_hazelnut} shows improved localization from both OrthLatt-UwU and SBE-OrthLatt-UwU. Notably, only SBE-OrthLatt-UwU improved over the baseline in both tasks, highlighting its balanced performance and superior localization.
\begin{table}[t!]
\caption{\justifying{Segmentation and Detection AUROCs of CFLOW-AD pipeline with the best SBE-OrthLatt-UwU, OrthLatt-UwU WaveCNet, and baseline ResNet18 encoders for hazelnut category in MVTec AD.\\}}
\centering
\scalebox{1.0}{
\begin{tabular}{ccc}
\hline
\multicolumn{1}{|l|}{\textbf{Models}} & \multicolumn{1}{|c|}{\textbf{SegAUROC}} & \multicolumn{1}{|c|}{\textbf{DetAUROC}}\\ \hline
\multicolumn{1}{|l|}{Baseline\cite{OrthLatt_UwU}} & \multicolumn{1}{|c|}{96.45$\%$} & \multicolumn{1}{|c|}{92.46$\%$}\\ \hline
\multicolumn{1}{|l|}{SBE-OrthLatt-UwU DB2} & \multicolumn{1}{|c|}{96.99$\%$} & \multicolumn{1}{|c|}{94.46$\%$}\\ \hline
\multicolumn{1}{|l|}{OrthLatt-UwU DB2\cite{OrthLatt_UwU}} & \multicolumn{1}{|c|}{97.15$\%$} & \multicolumn{1}{|c|}{89.57$\%$}\\ \hline
\multicolumn{1}{|l|}{WaveCNet Sym4\cite{OrthLatt_UwU}} & \multicolumn{1}{|c|}{96.20$\%$} & \multicolumn{1}{|c|}{87.61$\%$}\\ \hline
\end{tabular}}
\label{table:MVTecAD_hazelnut}
\end{table}
\section{Conclusion}
\label{sec:conclude}
This study introduces a stop-band energy (SBE) constraint to enhance frequency selectivity in tunable wavelets, improving CNN performance in high-frequency tasks like image classification and anomaly detection. Applied to OrthLatt-UwU in ResNet, the SBE constraint boosts performance on texture-rich datasets while maintaining competitive results on low-resolution ones. Though not always improving top-1 accuracy, it enhances the model’s ability to capture fine details. The approach is computationally efficient and broadly applicable, with future extensions planned for Transformer architectures.
\bibliographystyle{ieeetr}
\bibliography{refs}

\end{document}